\documentclass[sigconf,nonacm,twoside]{acmart} 
\usepackage{booktabs}
\usepackage{siunitx}
\usepackage{comment}
\usepackage{subcaption}
\usepackage{makecell} 
\usepackage[table]{xcolor} 

\AtBeginDocument{%
  }

\setcopyright{none}
\copyrightyear{2025}
\acmYear{2025} 
\acmDOI{}
\acmConference[HCAI Workshop @ CIKM '25]{Human-Centric AI Workshop}{November 14, 2025}{Seoul, Republic of Korea}

\begin{document}

\title{Language over Content: Tracing Cultural Understanding in Multilingual Large Language Models}
\renewcommand{\shorttitle}{Language over Content: Tracing Cultural Understanding in Multilingual Large Language Models}

\author{Seungho Cho}
\affiliation{%
  \institution{KAIST}
  \city{Daejeon}
  \country{Republic of Korea}}
\email{cho.seungho@kaist.ac.kr}

\author{Changgeon Ko}
\affiliation{%
  \institution{KAIST}
  \city{Daejeon}
  \country{Republic of Korea}}
\email{pencaty@kaist.ac.kr}

\author{Eui Jun Hwang}
\affiliation{%
  \institution{KAIST}
  \city{Daejeon}
  \country{Republic of Korea}}
\email{ehwa20@kaist.ac.kr}

\author{Junmyeong Lee}
\affiliation{%
  \institution{KAIST}
  \city{Daejeon}
  \country{Republic of Korea}}
\email{david516@kaist.ac.kr}

\author{Huije Lee}
\affiliation{%
  \institution{KAIST}
  \city{Daejeon}
  \country{Republic of Korea}}
\email{huijelee@kaist.ac.kr}

\author{Jong C. Park}
\affiliation{%
  \institution{KAIST}
  \city{Daejeon}
  \country{Republic of Korea}}
\email{jongpark@kaist.ac.kr}
\authornote{Corresponding author.}

\renewcommand{\shortauthors}{Cho et al.}


\begin{abstract}
Large language models (LLMs) are increasingly used across diverse cultural contexts, making accurate cultural understanding essential. Prior evaluations have mostly focused on output-level performance, obscuring the factors that drive differences in responses, while studies using circuit analysis have covered few languages and rarely focused on culture. In this work, we trace LLMs’ internal cultural understanding mechanisms by measuring activation path overlaps when answering semantically equivalent questions under two conditions: varying the target country while fixing the question language, and varying the question language while fixing the country. We also use same-language country pairs to disentangle language from cultural aspects. Results show that internal paths overlap more for same-language, cross-country questions than for cross-language, same-country questions, indicating strong language-specific patterns. Notably, the South Korea–North Korea pair exhibits low overlap and high variability, showing that linguistic similarity does not guarantee aligned internal representation.
\end{abstract}



\maketitle
\fancypagestyle{firstpage}{
\fancyhf{}  
\fancyfoot[C]{\thepage} 
\fancyfoot[L]{\footnotesize © 2025 Copyright held by the owner/author(s).}
\renewcommand{\headrulewidth}{0pt}
 \renewcommand{\footrulewidth}{0pt}
}
\makeatletter
\setlength{\footskip}{30pt}
\fancypagestyle{plain}{%
  \fancyfoot{}
  \fancyhead[LE]{\footnotesize HCAI Workshop @ CIKM ’25, November 14, 2025, Seoul, Republic of Korea}
  \fancyhead[RO]{\footnotesize HCAI Workshop @ CIKM ’25, November 14, 2025, Seoul, Republic of Korea}
  \fancyfoot[C]{\thepage} 
  \renewcommand{\footrulewidth}{0pt} 
}
\pagestyle{plain}
\thispagestyle{plain}
\makeatother
\thispagestyle{firstpage}

\section{Introduction}
\label{1_intro}

\begin{figure}[t]
  \centering
  \includegraphics[width=0.45\textwidth]{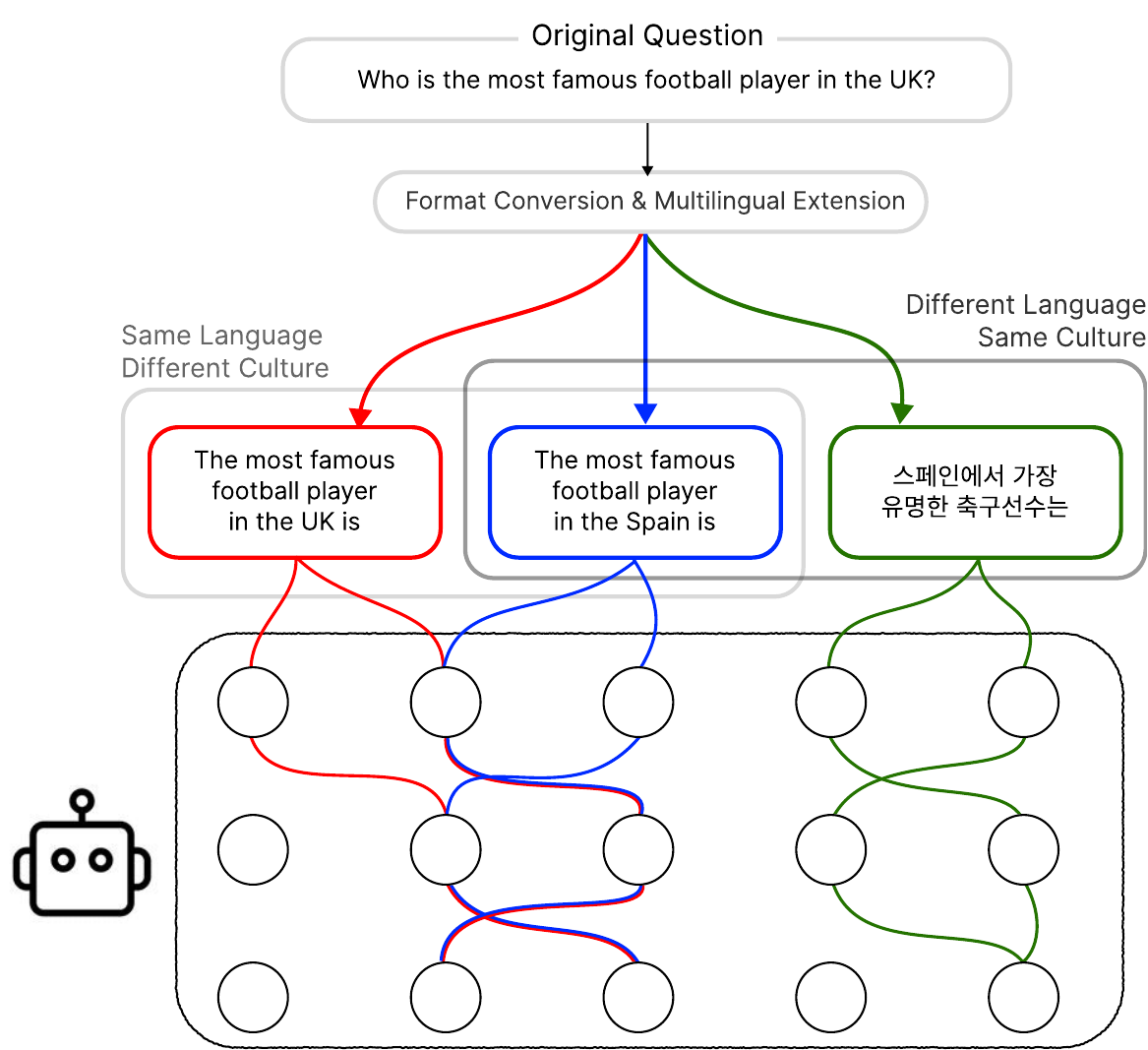}
  \vspace{-0.5em}
  \captionsetup{skip=7pt}
  \caption{Overview of Tracing Cultural Understanding in Multilingual Large Language Models.}
  \label{fig:fig1}
  \vspace{-1.5em}
\end{figure}

Large language models (LLMs) have achieved strong performance across a wide range of tasks, including translation, reasoning, and question answering~\cite{openai2025gptoss120bgptoss20bmodel,kimiteam2025kimik2openagentic}. However, when it comes to culture, responses can vary to the speaker, country, or region, since cultural knowledge reflects social norms, historical events, and linguistic nuances that differ across societies and evolve over time~\cite{hershcovich-etal-2022-challenges,myung2024blend}. Therefore, it is crucial for LLMs to develop a robust understanding of diverse cultural contexts in order to provide reliable and contextually appropriate outputs~\cite{liu-etal-2025-culturally,salemi2023lamp}.

Since LLMs acquire cultural knowledge from diverse language sources during pretraining~\cite{zhao-etal-2024-tracing,goldman2025eclekticnovelchallengeset,zhang-etal-2025-cross}, linguistic and cultural signals are often intertwined, making it necessary to study them together rather than in isolation~\cite{hershcovich-etal-2022-challenges}. Nevertheless, most evaluations of cultural understanding have focused only on final outputs, which obscures the factors driving differences in responses, such as question language, cultural knowledge, or their interaction~\cite{li2024culturegen,myung2024blend,singh-etal-2025-global}. Although some research has attempted to reveal the internal circuits of multilingual LLMs by varying question language~\cite{lindsey2025biology,zhang2025the,zhao2024how}, little work has examined how models represent and process cultural knowledge within multilingual contexts~\cite{zhang2025the,resck_explainability_2025}.

In this study, we address these gaps by tracing how LLMs internally use their cultural understandings. Specifically, we measure how models’ internal paths change when answering semantically equivalent cultural questions under two conditions: (1) varying the target country while fixing the question language, and (2) varying the language of the question while fixing the target country. To further disentangle cultural and linguistic aspects, we include special country pairs such as South Korea–North Korea, the US–UK, and Spain-Mexico. These pairs share similar or identical languages but differ culturally, allowing us to better separate language-driven from culture-driven signals. This design enables us to ask whether linguistic cues dominate cultural knowledge representation or whether the two interact in more complex ways.

Our experiments show that internal path overlap is greater for same-language, cross-country questions than for cross-language, same-country questions, indicating a strong language-specific pattern in representing cultural knowledge. Notably, the South Korea–North Korea pair shows unusually low overlaps and high variability across question languages compared with other same-language pairs, highlighting the need for further analysis. Overall, these findings suggest that multilingual LLMs rely heavily on language-specific circuits when representing and applying cultural knowledge.

Our contributions and findings can be summarized as follows:
\begin{itemize}
    \item We highlight the need to investigate how multilingual LLMs internally represent cultural understanding.
    \item Our experiments show that internal paths overlap more when questions are in similar languages across cultures than when they are in different languages for the same culture, indicating a strong language-specific pattern.
    \item We find cases where models exhibit distinct internal patterns despite high linguistic similarity, highlighting the need for further investigation.
    
\end{itemize}

\section{Related Work}
\label{2_related_work}

\subsection{Mechanistic Interpretability of LLMs}
Recent studies have actively investigated methods to understand the internal mechanisms of LLMs by first identifying interpretable features in their computations and then constructing circuits to capture how these features interact~\cite{ameisen2025circuit,miller2024transformer,zhang2025the,zhao2024how}. Some early works directly treated raw neurons as interpretable features~\cite{tang-etal-2024-language,kojima-etal-2024-multilingual,ying-etal-2025-disentangling}; however, the polysemantic nature of neurons made it difficult to derive clear interpretations of model behavior~\cite{elhage2022superposition}. To address this limitation, subsequent studies trained sparse coding models such as SAE to decompose MLP representations into interpretable features~\cite{bricken2023monosemanticity,DBLP:journals/corr/abs-2503-05613}, but these approaches lacked input invariance, preventing general conclusions about model behavior. Transcoder overcomes these challenges by decomposing MLP computations and enabling input-invariant, feature-level circuit analysis~\cite{dunefsky2024transcoders,ameisen2025circuit}. This method allows for the extraction of more general interpretable features and the direct computation of feature interactions, facilitating a detailed analysis of internal model flows. Using this approach, we construct circuits to study the internal mechanisms that occur during the LLM’s answer generation process.

\subsection{Cultural Understanding of Multilingual LLMs}
Recent studies have sought to evaluate the cultural understanding of multilingual LLMs, leading to the development of benchmarks that incorporate locally collected, culture-specific data. While these benchmarks better reflect target cultures, they remain limited in scope, particularly in the number of languages considered and their ability to capture multilingual usage scenarios. Moreover, most prior studies have assessed model performance only at the output level~\cite{li2024culturegen,myung2024blend,singh-etal-2025-global}, leaving the internal mechanisms underlying cultural understanding largely unexplored. Although recent research has begun to address interpretability, it has primarily focused on multilinguality rather than culture~\cite{zhang2025the,resck_explainability_2025,zhao2024how}, with analyses often restricted to language-specific neurons and narrow language coverage~\cite{ying-etal-2025-disentangling}. In this work, we extend this line of research by examining internal circuits when multilingual LLMs answer culturally relevant questions across more diverse multilingual settings.

\section{Tracing Knowledge Circuit}
\label{3_method}

\subsection{Task Formulation}

We define $Q_{L,C}$ as the set of culture-related questions asked in language $L$ about country $C$. Correspondingly, $P(Q_{L,C})$ represents the internal activation paths within the LLM when answering the questions in $Q_{L,C}$. We followed the approach in ~\cite{dunefsky2024transcoders} to extract interpretable features (nodes), to measure attributions (edges) and to construct the circuits (internal path).

To analyze the interplay of language and culture in the model’s internal processing, we measure the overlap between activation paths under two cases.
First, we fix the language $L$ and compare the similarity between internal paths activated by questions about two different countries $C_n$ and $C_m$, denoted as $Sim(P(Q_{L,C_n}), P(Q_{L,C_m}))$. 
Second, we fix the country $C$ and compare internal path similarity for questions asked in two different languages $L_n$ and $L_m$, denoted as $Sim(P(Q_{L_n,C}), P(Q_{L_m,C}))$. By contrasting these overlaps, we assess whether the model’s cultural knowledge representation is predominately influenced by the input language or the cultural content itself.

The internal paths $P(Q_{L,C})$ are extracted as weighted subgraphs that represent the model’s internal feature attributions during answer generation. Nodes correspond to interpretable features, while edges capture attribution strength between features. We normalize each edge’s weight so that the sum of absolute attributions equals one. We then quantify path similarity using Weighted Jaccard Similarity~\cite{wiki:Jaccard_index}, treating missing edges as having zero weight. Similarity scores close to 1 indicate largely overlapping internal processing paths, whereas scores near 0 indicate distinct mechanisms.

\subsection{Data Construction}

We construct $Q_{L,C}$ using the culture-specific benchmark dataset BLEnD~\cite{myung2024blend}. 
From the question set, we randomly select 50 questions, ensuring minimal semantic overlap, to create our experimental dataset.

\subsubsection{Country and Language Selection}

For cross-cultural analysis, we select seven countries: South Korea (KR), North Korea (KP), the United States (US), the United Kingdom (UK), Spain (ES), Mexico (MX), and China (CN). 
To study cases where language is shared but cultural contexts vary, we include three pairs of linguistically related countries: South Korea–North Korea, Mexico–Spain, and the United Kingdom–United States. 
These pairings minimize linguistic variation while emphasizing cultural differences. 
As a contrasting case, we add China to represent a distinct language group. 
While languages within each pair are highly similar, subtle distinctions in vocabulary and grammar remain. 
For this reason, we treat them as separate languages throughout our analysis.

\subsubsection{Question Format Conversion}

To facilitate next-token prediction and simplify the analysis of the model’s internal representations, we convert interrogative questions into declarative statements. 
This conversion enables the model to generate answers as continuations within a unified framework, providing inputs that more closely match its training distribution~\cite{mitchell-etal-2022-enhancing}. For example, the question ``Who is the most famous football player in the UK?'' can be converted as ``The most famous football player in the UK is\_''. 
This process preserves grammatical correctness and natural word order in each language, inserting spaces where required for accurate token generation (except for Chinese, which does not use spaces).  

\subsubsection{Multilingual Extension}

The original BLEnD dataset provides each cultural question set only in its corresponding language.  
To extend coverage, we translate each question set into all languages used in our experiments, creating 49 $Q_{L,C}$ question sets.
For example, a question about the most famous football player in South Korea is expressed not only in Korean, but also in English, Chinese, and Spanish.  
This design enables systematic analysis of how language inputs and culture jointly influence the model's internal path selection.

\subsection{Implementation Details}

For questions related to cultural knowledge, we used the Gemma 2\footnote{https://huggingface.co/google/gemma-2-2b}~\cite{DBLP:journals/corr/abs-2408-00118}. To avoid potential effects on the model’s internal paths, we employed the base version rather than the instruction-tuned model. For internal path extraction, we used Gemma Scope Transcoder\footnote{https://huggingface.co/google/gemma-scope-2b-pt-transcoders}~\cite{lieberum-etal-2024-gemma,dunefsky2024transcoders} to obtain interpretable features. In the dataset reformulation and extension steps, the questions were generated with GPT-4o and verified with o4-mini via the OpenAI API\footnote{https://platform.openai.com/}. 

\section{Analysis}
\label{4_result}

\begin{table}[ht!]
    \centering
    \small
    \caption{(a) Path overlap across target cultures with the question language fixed. (b) Path overlap across question languages with the target culture fixed.}
    \vspace{-1.0em}
    \label{tab:main}
    \begin{minipage}[t]{0.48\textwidth}
    \begin{subtable}{\linewidth}
        \centering
        \caption{Fixed Language}
        \vspace{-0.1em}
        \label{tab:fixed_lang}
        \begin{tabular}{cc|ccccccc}
        \toprule
        \multicolumn{2}{c}{} & \multicolumn{7}{c}{$\mathbf{L}$} \\
        \cmidrule(lr){3-9}
        $\mathbf{C}_1$ & $\mathbf{C}_2$ & {KR} & {KP} & {US} & {UK} & {ES} & {MX} & {CN} \\
        \midrule
        KR & KP & \cellcolor[HTML]{FEF8F8}0.10 & \cellcolor[HTML]{FEFCFC}0.04 & \cellcolor[HTML]{F9E2DF}0.44 & \cellcolor[HTML]{F9E0DE}0.46 & \cellcolor[HTML]{F8D9D6}0.57 & \cellcolor[HTML]{F8D9D7}0.56 & \cellcolor[HTML]{F8DCDA}0.52\\ KR & US & \cellcolor[HTML]{FEFCFB}0.05 & \cellcolor[HTML]{FEFCFC}0.04 & \cellcolor[HTML]{FAE6E4}0.37 & \cellcolor[HTML]{FAE5E3}0.39 & \cellcolor[HTML]{FDF4F3}0.16 & \cellcolor[HTML]{FDF4F3}0.16 & \cellcolor[HTML]{FCF1F0}0.21\\ KR & UK & \cellcolor[HTML]{FEFBFB}0.06 & \cellcolor[HTML]{FEFCFC}0.04 & \cellcolor[HTML]{FAE6E4}0.37 & \cellcolor[HTML]{FAE6E4}0.38 & \cellcolor[HTML]{FAE4E2}0.40 & \cellcolor[HTML]{FAE3E1}0.42 & \cellcolor[HTML]{F9E1DF}0.45\\ KR & ES & \cellcolor[HTML]{FBEAE9}0.31 & \cellcolor[HTML]{FEFCFC}0.04 & \cellcolor[HTML]{FDF4F3}0.17 & \cellcolor[HTML]{FDF4F3}0.17 & \cellcolor[HTML]{FDF5F4}0.15 & \cellcolor[HTML]{FDF5F4}0.15 & \cellcolor[HTML]{F9E2DF}0.44\\ KR & MX & \cellcolor[HTML]{FBEBE9}0.30 & \cellcolor[HTML]{FAE6E4}0.38 & \cellcolor[HTML]{FDF4F3}0.17 & \cellcolor[HTML]{FDF4F3}0.17 & \cellcolor[HTML]{FDF4F3}0.16 & \cellcolor[HTML]{FDF5F4}0.15 & \cellcolor[HTML]{F9E2E0}0.43\\ KR & CN & \cellcolor[HTML]{FEFBFB}0.06 & \cellcolor[HTML]{FEFCFC}0.04 & \cellcolor[HTML]{FDF4F3}0.17 & \cellcolor[HTML]{FDF4F3}0.17 & \cellcolor[HTML]{FDF4F3}0.16 & \cellcolor[HTML]{FDF4F3}0.16 & \cellcolor[HTML]{FDF4F3}0.16\\ \midrule
        KP & US & \cellcolor[HTML]{FAE8E6}0.35 & \cellcolor[HTML]{FAE8E6}0.35 & \cellcolor[HTML]{FBEAE9}0.31 & \cellcolor[HTML]{FBEAE9}0.31 & \cellcolor[HTML]{FDF5F4}0.15 & \cellcolor[HTML]{FDF5F4}0.15 & \cellcolor[HTML]{FDF2F1}0.19\\ KP & UK & \cellcolor[HTML]{FAE7E5}0.36 & \cellcolor[HTML]{FAE7E5}0.36 & \cellcolor[HTML]{FBEBE9}0.30 & \cellcolor[HTML]{FBEBE9}0.30 & \cellcolor[HTML]{FAE7E5}0.36 & \cellcolor[HTML]{FAE6E4}0.37 & \cellcolor[HTML]{FAE3E1}0.42\\ KP & ES & \cellcolor[HTML]{FEFCFC}0.04 & \cellcolor[HTML]{FEFCFC}0.04 & \cellcolor[HTML]{FDF4F3}0.16 & \cellcolor[HTML]{FDF4F3}0.16 & \cellcolor[HTML]{FDF6F5}0.14 & \cellcolor[HTML]{FDF6F5}0.14 & \cellcolor[HTML]{F9E2E0}0.43\\ KP & MX & \cellcolor[HTML]{FEFCFC}0.04 & \cellcolor[HTML]{FEFCFC}0.04 & \cellcolor[HTML]{FDF5F4}0.15 & \cellcolor[HTML]{FDF4F3}0.16 & \cellcolor[HTML]{FDF6F5}0.14 & \cellcolor[HTML]{FDF6F5}0.14 & \cellcolor[HTML]{F9E2DF}0.44\\ KP & CN & \cellcolor[HTML]{FAE6E4}0.38 & \cellcolor[HTML]{FAE5E3}0.39 & \cellcolor[HTML]{FDF4F3}0.16 & \cellcolor[HTML]{FDF4F3}0.16 & \cellcolor[HTML]{FDF5F4}0.15 & \cellcolor[HTML]{FDF5F4}0.15 & \cellcolor[HTML]{FDF4F3}0.16\\ \midrule
        US & UK & \cellcolor[HTML]{F7D7D5}0.59 & \cellcolor[HTML]{F7D8D5}0.58 & \cellcolor[HTML]{F8D9D7}0.56 & \cellcolor[HTML]{F8D9D7}0.56 & \cellcolor[HTML]{FBEDEC}0.27 & \cellcolor[HTML]{FBEDEC}0.27 & \cellcolor[HTML]{FBEDEC}0.27\\ US & ES & \cellcolor[HTML]{FEFCFB}0.05 & \cellcolor[HTML]{FEFCFC}0.04 & \cellcolor[HTML]{FDF2F1}0.19 & \cellcolor[HTML]{FDF2F1}0.19 & \cellcolor[HTML]{FDF4F3}0.17 & \cellcolor[HTML]{FDF4F3}0.16 & \cellcolor[HTML]{FCF2F1}0.20\\ US & MX & \cellcolor[HTML]{FEFCFC}0.04 & \cellcolor[HTML]{FEFCFB}0.05 & \cellcolor[HTML]{FDF2F1}0.19 & \cellcolor[HTML]{FDF2F1}0.19 & \cellcolor[HTML]{FDF4F3}0.17 & \cellcolor[HTML]{FDF4F3}0.16 & \cellcolor[HTML]{FCF1F0}0.21\\ US & CN & \cellcolor[HTML]{F8DEDB}0.50 & \cellcolor[HTML]{F9DFDC}0.48 & \cellcolor[HTML]{FDF2F1}0.19 & \cellcolor[HTML]{FDF2F1}0.19 & \cellcolor[HTML]{FDF3F2}0.18 & \cellcolor[HTML]{FDF4F3}0.17 & \cellcolor[HTML]{FAE5E3}0.39\\ \midrule
        UK & ES & \cellcolor[HTML]{FEFCFB}0.05 & \cellcolor[HTML]{FEFCFC}0.04 & \cellcolor[HTML]{FDF2F1}0.19 & \cellcolor[HTML]{FDF3F2}0.18 & \cellcolor[HTML]{FDF4F3}0.16 & \cellcolor[HTML]{FDF4F3}0.16 & \cellcolor[HTML]{F8DEDB}0.50\\ UK & MX & \cellcolor[HTML]{FEFCFC}0.04 & \cellcolor[HTML]{FEFCFC}0.04 & \cellcolor[HTML]{FDF3F2}0.18 & \cellcolor[HTML]{FDF3F2}0.18 & \cellcolor[HTML]{FDF4F3}0.16 & \cellcolor[HTML]{FDF4F3}0.16 & \cellcolor[HTML]{F9E0DD}0.47\\ UK & CN & \cellcolor[HTML]{F8DDDA}0.51 & \cellcolor[HTML]{F8DDDA}0.51 & \cellcolor[HTML]{FDF2F1}0.19 & \cellcolor[HTML]{FDF3F2}0.18 & \cellcolor[HTML]{FDF4F3}0.16 & \cellcolor[HTML]{FDF4F3}0.16 & \cellcolor[HTML]{FDF4F3}0.17\\ \midrule
        ES & MX & \cellcolor[HTML]{FAE4E1}0.41 & \cellcolor[HTML]{FEFCFB}0.05 & \cellcolor[HTML]{F8DBD8}0.54 & \cellcolor[HTML]{F8DBD9}0.53 & \cellcolor[HTML]{F7D7D4}0.60 & \cellcolor[HTML]{F7D7D5}0.59 & \cellcolor[HTML]{F8D9D7}0.56\\ ES & CN & \cellcolor[HTML]{FEFCFB}0.05 & \cellcolor[HTML]{FEFCFC}0.04 & \cellcolor[HTML]{F9E2E0}0.43 & \cellcolor[HTML]{F9E2DF}0.44 & \cellcolor[HTML]{F9DEDC}0.49 & \cellcolor[HTML]{F8DDDA}0.51 & \cellcolor[HTML]{FDF6F5}0.14\\ \midrule
        MX & CN & \cellcolor[HTML]{FEFCFC}0.04 & \cellcolor[HTML]{FEFCFB}0.05 & \cellcolor[HTML]{F9E2DF}0.44 & \cellcolor[HTML]{F9E2E0}0.43 & \cellcolor[HTML]{F8DDDA}0.51 & \cellcolor[HTML]{F8DCDA}0.52 & \cellcolor[HTML]{FDF5F4}0.15\\
        \bottomrule
        \end{tabular}
    \end{subtable}

    \vspace{1.0em}

    \begin{subtable}{\linewidth}
        \centering
        \caption{Fixed Culture}
        \label{tab:fixed_culture}
        \vspace{-0.1em}
        \begin{tabular}{cc|ccccccc}
        \toprule
        \multicolumn{2}{c}{} & \multicolumn{7}{c}{$\mathbf{C}$} \\
        \cmidrule(lr){3-9}
        $\mathbf{L}_1$ & $\mathbf{L}_2$ & {KR} & {KP} & {US} & {UK} & {ES} & {MX} & {CN} \\
        \midrule
        KR & KP & \cellcolor[HTML]{FDF4F3}0.16 & \cellcolor[HTML]{FAE6E4}0.38 & \cellcolor[HTML]{FAE7E5}0.36 & \cellcolor[HTML]{FAE6E4}0.37 & \cellcolor[HTML]{FEFBFB}0.06 & \cellcolor[HTML]{FDF5F4}0.15 & \cellcolor[HTML]{FAE6E4}0.38 \\ KR & US & \cellcolor[HTML]{FFFEFE}0.01 & \cellcolor[HTML]{FFFEFE}0.01 & \cellcolor[HTML]{FFFEFE}0.02 & \cellcolor[HTML]{FFFEFE}0.02 & \cellcolor[HTML]{FFFEFE}0.01 & \cellcolor[HTML]{FFFEFE}0.01 & \cellcolor[HTML]{FFFEFE}0.01 \\ KR & UK & \cellcolor[HTML]{FFFEFE}0.01 & \cellcolor[HTML]{FFFEFE}0.01 & \cellcolor[HTML]{FFFEFE}0.02 & \cellcolor[HTML]{FFFEFE}0.02 & \cellcolor[HTML]{FFFEFE}0.01 & \cellcolor[HTML]{FFFEFE}0.01 & \cellcolor[HTML]{FFFEFE}0.01 \\ KR & ES & \cellcolor[HTML]{FFFEFE}0.01 & \cellcolor[HTML]{FFFEFE}0.01 & \cellcolor[HTML]{FFFEFE}0.01 & \cellcolor[HTML]{FFFEFE}0.02 & \cellcolor[HTML]{FFFEFE}0.01 & \cellcolor[HTML]{FFFEFE}0.01 & \cellcolor[HTML]{FFFEFE}0.02 \\ KR & MX & \cellcolor[HTML]{FFFEFE}0.01 & \cellcolor[HTML]{FFFEFE}0.01 & \cellcolor[HTML]{FFFEFE}0.01 & \cellcolor[HTML]{FFFEFE}0.02 & \cellcolor[HTML]{FFFEFE}0.01 & \cellcolor[HTML]{FFFEFE}0.01 & \cellcolor[HTML]{FFFEFE}0.02 \\ KR & CN & \cellcolor[HTML]{FFFEFE}0.02 & \cellcolor[HTML]{FFFDFD}0.03 & \cellcolor[HTML]{FFFEFE}0.02 & \cellcolor[HTML]{FFFDFD}0.03 & \cellcolor[HTML]{FFFEFE}0.02 & \cellcolor[HTML]{FFFEFE}0.02 & \cellcolor[HTML]{FFFEFE}0.02 \\ \midrule KP & US & \cellcolor[HTML]{FFFEFE}0.01 & \cellcolor[HTML]{FFFEFE}0.01 & \cellcolor[HTML]{FFFEFE}0.02 & \cellcolor[HTML]{FFFEFE}0.02 & \cellcolor[HTML]{FFFEFE}0.01 & \cellcolor[HTML]{FFFEFE}0.01 & \cellcolor[HTML]{FFFEFE}0.02 \\ KP & UK & \cellcolor[HTML]{FFFEFE}0.01 & \cellcolor[HTML]{FFFEFE}0.01 & \cellcolor[HTML]{FFFEFE}0.02 & \cellcolor[HTML]{FFFEFE}0.02 & \cellcolor[HTML]{FFFEFE}0.01 & \cellcolor[HTML]{FFFEFE}0.01 & \cellcolor[HTML]{FFFEFE}0.02 \\ KP & ES & \cellcolor[HTML]{FFFEFE}0.01 & \cellcolor[HTML]{FFFEFE}0.01 & \cellcolor[HTML]{FFFEFE}0.01 & \cellcolor[HTML]{FFFEFE}0.02 & \cellcolor[HTML]{FFFEFE}0.01 & \cellcolor[HTML]{FFFEFE}0.01 & \cellcolor[HTML]{FFFEFE}0.02 \\ KP & MX & \cellcolor[HTML]{FFFEFE}0.01 & \cellcolor[HTML]{FFFEFE}0.01 & \cellcolor[HTML]{FFFEFE}0.01 & \cellcolor[HTML]{FFFEFE}0.02 & \cellcolor[HTML]{FFFEFE}0.01 & \cellcolor[HTML]{FFFEFE}0.01 & \cellcolor[HTML]{FFFEFE}0.02 \\ KP & CN & \cellcolor[HTML]{FFFEFE}0.02 & \cellcolor[HTML]{FFFDFD}0.03 & \cellcolor[HTML]{FFFEFE}0.02 & \cellcolor[HTML]{FFFEFE}0.02 & \cellcolor[HTML]{FFFEFE}0.02 & \cellcolor[HTML]{FFFEFE}0.02 & \cellcolor[HTML]{FFFEFE}0.02 \\ \midrule US & UK & \cellcolor[HTML]{F3C2BD}0.91 & \cellcolor[HTML]{F3C0BB}0.94 & \cellcolor[HTML]{F3BFBB}0.95 & \cellcolor[HTML]{F3C1BC}0.93 & \cellcolor[HTML]{F2BEB9}0.97 & \cellcolor[HTML]{F3BFBB}0.95 & \cellcolor[HTML]{F3C3BF}0.89 \\ US & ES & \cellcolor[HTML]{FFFEFE}0.02 & \cellcolor[HTML]{FFFEFE}0.02 & \cellcolor[HTML]{FFFEFE}0.02 & \cellcolor[HTML]{FFFEFE}0.02 & \cellcolor[HTML]{FFFDFD}0.03 & \cellcolor[HTML]{FFFDFD}0.03 & \cellcolor[HTML]{FFFDFD}0.03 \\ US & MX & \cellcolor[HTML]{FFFEFE}0.02 & \cellcolor[HTML]{FFFEFE}0.02 & \cellcolor[HTML]{FFFEFE}0.02 & \cellcolor[HTML]{FFFEFE}0.02 & \cellcolor[HTML]{FFFEFE}0.02 & \cellcolor[HTML]{FFFDFD}0.03 & \cellcolor[HTML]{FFFDFD}0.03 \\ US & CN & \cellcolor[HTML]{FFFEFE}0.02 & \cellcolor[HTML]{FFFEFE}0.02 & \cellcolor[HTML]{FFFEFE}0.02 & \cellcolor[HTML]{FFFEFE}0.02 & \cellcolor[HTML]{FFFEFE}0.02 & \cellcolor[HTML]{FFFEFE}0.02 & \cellcolor[HTML]{FFFEFE}0.02 \\ \midrule UK & ES & \cellcolor[HTML]{FFFEFE}0.02 & \cellcolor[HTML]{FFFEFE}0.02 & \cellcolor[HTML]{FFFEFE}0.02 & \cellcolor[HTML]{FFFEFE}0.02 & \cellcolor[HTML]{FFFDFD}0.03 & \cellcolor[HTML]{FFFDFD}0.03 & \cellcolor[HTML]{FFFDFD}0.03 \\ UK & MX & \cellcolor[HTML]{FFFEFE}0.02 & \cellcolor[HTML]{FFFEFE}0.02 & \cellcolor[HTML]{FFFEFE}0.02 & \cellcolor[HTML]{FFFEFE}0.02 & \cellcolor[HTML]{FFFEFE}0.02 & \cellcolor[HTML]{FFFDFD}0.03 & \cellcolor[HTML]{FFFDFD}0.03 \\ UK & CN & \cellcolor[HTML]{FFFEFE}0.02 & \cellcolor[HTML]{FFFEFE}0.02 & \cellcolor[HTML]{FFFEFE}0.02 & \cellcolor[HTML]{FFFEFE}0.02 & \cellcolor[HTML]{FFFEFE}0.02 & \cellcolor[HTML]{FFFEFE}0.02 & \cellcolor[HTML]{FFFEFE}0.02 \\ \midrule ES & MX & \cellcolor[HTML]{F6D3CF}0.66 & \cellcolor[HTML]{F6D0CD}0.70 & \cellcolor[HTML]{F6D3CF}0.66 & \cellcolor[HTML]{F6D2CF}0.67 & \cellcolor[HTML]{F6D1CD}0.69 & \cellcolor[HTML]{F6D3CF}0.66 & \cellcolor[HTML]{F6D1CD}0.69 \\ ES & CN & \cellcolor[HTML]{FFFEFE}0.01 & \cellcolor[HTML]{FFFEFE}0.02 & \cellcolor[HTML]{FFFEFE}0.01 & \cellcolor[HTML]{FFFEFE}0.01 & \cellcolor[HTML]{FFFEFE}0.02 & \cellcolor[HTML]{FFFEFE}0.02 & \cellcolor[HTML]{FFFEFE}0.01 \\ \midrule MX & CN & \cellcolor[HTML]{FFFEFE}0.01 & \cellcolor[HTML]{FFFEFE}0.01 & \cellcolor[HTML]{FFFEFE}0.01 & \cellcolor[HTML]{FFFEFE}0.01 & \cellcolor[HTML]{FFFEFE}0.02 & \cellcolor[HTML]{FFFEFE}0.02 & \cellcolor[HTML]{FFFEFE}0.01 \\
        \bottomrule
        \end{tabular}
    \end{subtable}
    \end{minipage}
    \vspace{-1.0em}
\end{table}

\begin{figure*}[t!]
\centering
\captionsetup[subfigure]{skip=0pt}
\captionsetup{skip=0pt}
    \begin{subfigure}[t]{0.49\textwidth}
        \centering
        \includegraphics[width=0.95\textwidth]{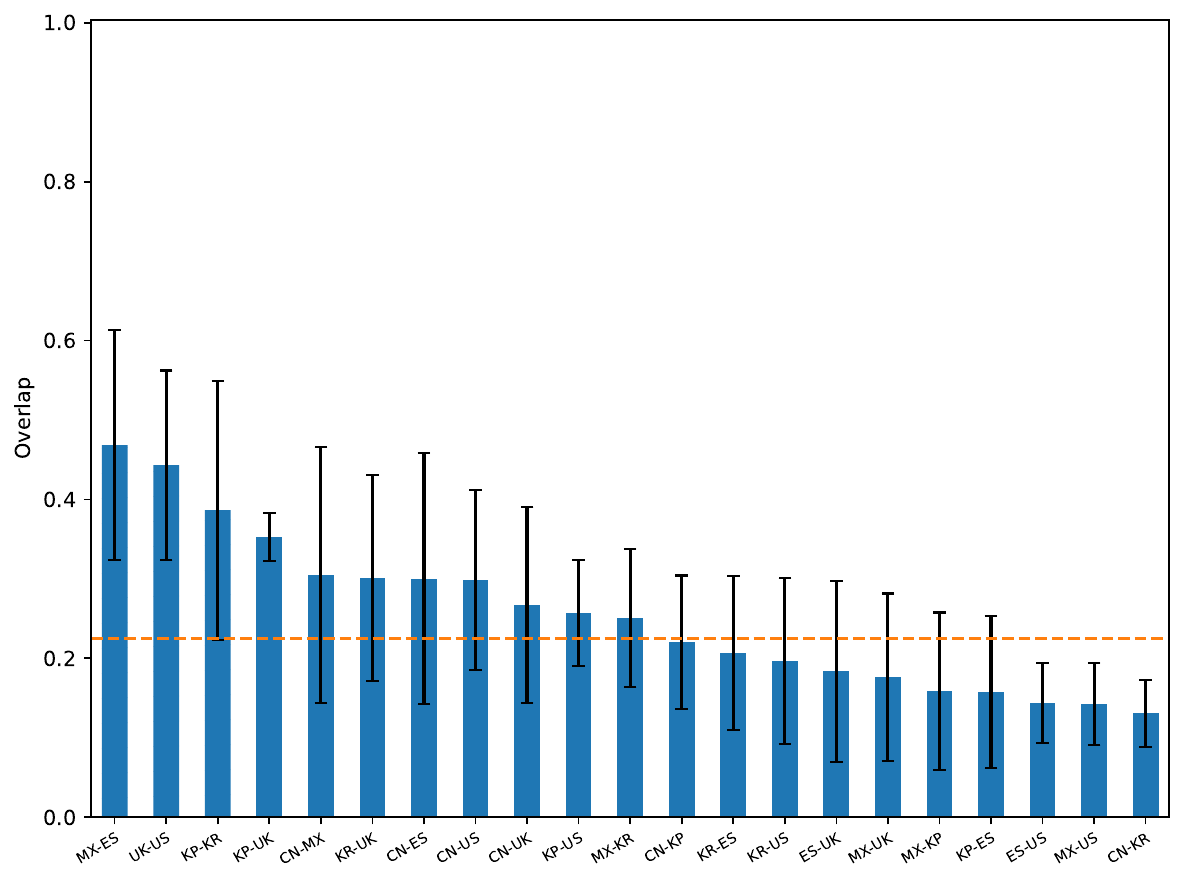}
        \caption{Fixed Language}
        \label{fig:2a}
    \end{subfigure}
\vspace{0.5em}
    \begin{subfigure}[t]{0.49\textwidth}
        \centering
        \includegraphics[width=0.95\textwidth]{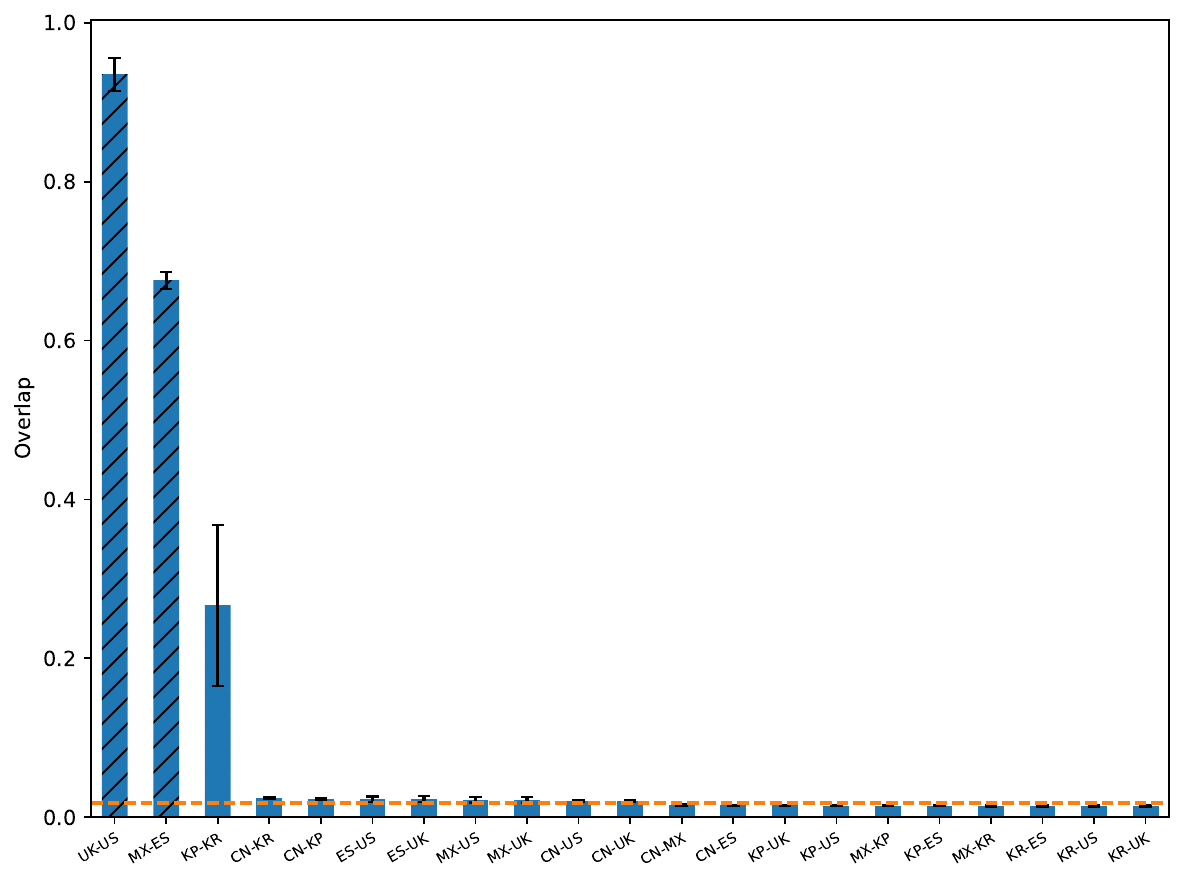}
        \caption{Fixed Culture}
        \label{fig:2b}
    \end{subfigure}
\vspace{-0.5em}

\caption{\small \textbf{(a)} Path overlap by country pair when the \emph{question language is fixed}; We find that overlaps remain relatively high, with linguistically similar country pairs showing especially high reuse of internal paths. \textbf{(b)} Path overlap by language pair when the \emph{target culture is fixed}; We find that overlaps drop markedly when the query language changes, indicating that language (rather than meaning) dominates internal path selection. Each bar shows the mean (±95\% CI), sorted in descending order; hatched bars denote linguistically similar pairs, and the orange horizontal line marks the overall average.}
\label{fig:2}
\vspace{-0.6em}
\end{figure*}

\subsection{Main Result}

Table~\ref{tab:main} presents similarity scores of internal path overlaps in the LLM under two conditions: fixed language and fixed culture. 
The results clearly show that the model’s internal path selection is much more affected by the language of the question than by the cultural context. 
When the question language is fixed (Table~\ref{tab:fixed_lang}), path overlap remains relatively high across different target cultures, especially among linguistically similar country pairs such as South Korea–North Korea, the United States–the United Kingdom, and Spain–Mexico. 
This suggests that language similarity strongly encourages reuse of internal paths.

Conversely, when the cultural context is fixed and the question language varies (Table~\ref{tab:fixed_culture}), path overlap drops significantly. 
This indicates that even semantically equivalent queries in different languages prompt the model to use markedly different internal paths. 
It implies that the model organizes and accesses cultural knowledge in a language-dependent manner, prioritizing linguistic form over semantic content when processing multilingual queries.

Figure~\ref{fig:2} summarizes these results by averaging scores for each pair, shown as descending bar charts with 95\% confidence intervals. Hatched bars highlight country pairs sharing similar languages, and the orange line marks the overall average. 
Figure~\ref{fig:2a} (fixed question language) highlights that linguistic similarity supports greater path overlap—possibly reflecting overlapping cultural traits as well. 
Figure~\ref{fig:2b} (fixed culture) shows low overlaps when question languages differ, supporting the conclusion that question language dominates internal path selection more than cultural context or semantic equivalence.

\begin{figure}[t]
\centering
\includegraphics[width=0.45\textwidth]{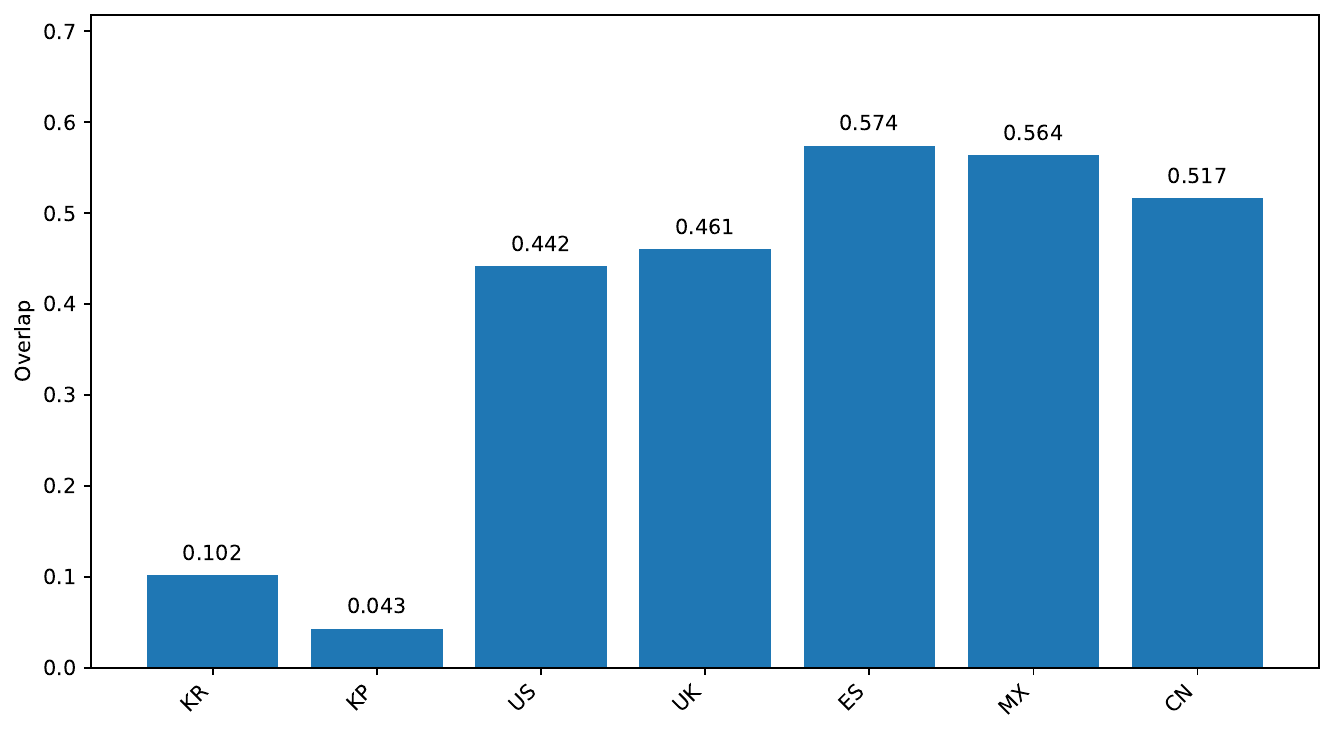}
\captionsetup{skip=0pt}
\caption{\small Path overlap between questions on South and North Korean culture by question language. We find that path overlaps are low in Korean languages than in non-Korean languages.}
\label{fig:3}
\vspace{-1.0em}
\end{figure}
\begin{figure}
\centering
\includegraphics[width=0.45\textwidth]{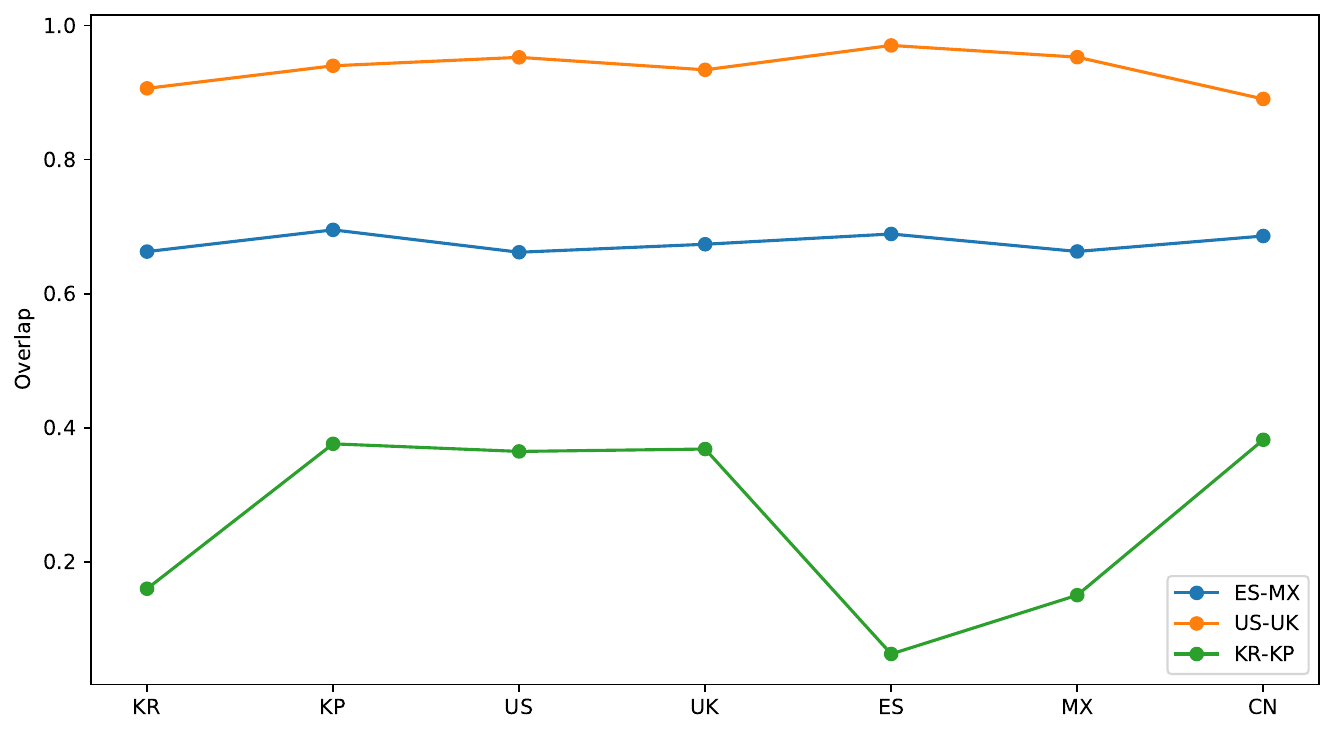}
\captionsetup{skip=0pt}
\caption{\small Path overlap between questions for similar-language pairs under a fixed target culture. US-UK and Spain-Mexico show high and stable overlap while South Korea-North Korea shows lower and more variable overlap.}
\label{fig:4}
\vspace{-1.5em}
\end{figure}

\subsection{Distinct Path Patterns in South and North Koreas}
South and North Koreas, despite sharing a similar language, exhibited distinct patterns compared with other linguistically similar pairs. Figure~\ref{fig:3} visualizes path overlaps for questions about South and North Korean culture across the various question languages. The results show higher overlaps in non-Korean languages and lower overlaps in Korean. Figure~\ref{fig:4} presents overlaps across similar-language pairs when the target culture is fixed. US–UK and Spain–Mexico maintained high and stable overlaps, while South Korea–North Korea showed lower and more variable overlaps. The reasons for these differences, whether due to linguistic, cultural, or other factors, remain unclear and warrant further investigation.

\section{Conclusion}
\label{5_conclusion}

We explored the internal mechanisms of cultural understanding in multilingual LLMs. To reflect realistic scenarios, we extended the cultural dataset to include multiple languages and measured the overlap of activated internal paths. Results show that query language affects internal path selection more strongly than target culture, and that cultural understanding is mainly stored in language-dependent paths. We also observed that politically unique contexts, such as South and North Koreas, are reflected in the model’s internal mechanisms. These findings offer key insights into how multilingual LLMs understand and utilize cultural knowledge.

Our analysis focused on internal path overlap, but interventions or circuit patching could clarify which features drive cultural knowledge processing and how language- and culture-related features interact. Some country pairs, such as South and North Koreas, show distinct patterns compared to other similar language pairs, yet the reasons remain unclear. Future work could further subdivide cultural knowledge into finer categories and include more languages, enabling a more comprehensive investigation of language–culture interactions within the model.


\bibliographystyle{ACM-Reference-Format}
\bibliography{references}

@misc{openai2025gptoss120bgptoss20bmodel,
      title={gpt-oss-120b \& gpt-oss-20b Model Card}, 
      author={OpenAI and Sandhini Agarwal and Lama Ahmad and Jason Ai and Sam Altman and Andy Applebaum and Edwin Arbus and Rahul K. Arora and Yu Bai and Bowen Baker and Haiming Bao and Boaz Barak and Ally Bennett and Tyler Bertao and Nivedita Brett and Eugene Brevdo and Greg Brockman and Sebastien Bubeck and Che Chang and Kai Chen and Mark Chen and Enoch Cheung and Aidan Clark and Dan Cook and Marat Dukhan and Casey Dvorak and Kevin Fives and Vlad Fomenko and Timur Garipov and Kristian Georgiev and Mia Glaese and Tarun Gogineni and Adam Goucher and Lukas Gross and Katia Gil Guzman and John Hallman and Jackie Hehir and Johannes Heidecke and Alec Helyar and Haitang Hu and Romain Huet and Jacob Huh and Saachi Jain and Zach Johnson and Chris Koch and Irina Kofman and Dominik Kundel and Jason Kwon and Volodymyr Kyrylov and Elaine Ya Le and Guillaume Leclerc and James Park Lennon and Scott Lessans and Mario Lezcano-Casado and Yuanzhi Li and Zhuohan Li and Ji Lin and Jordan Liss and Lily and Liu and Jiancheng Liu and Kevin Lu and Chris Lu and Zoran Martinovic and Lindsay McCallum and Josh McGrath and Scott McKinney and Aidan McLaughlin and Song Mei and Steve Mostovoy and Tong Mu and Gideon Myles and Alexander Neitz and Alex Nichol and Jakub Pachocki and Alex Paino and Dana Palmie and Ashley Pantuliano and Giambattista Parascandolo and Jongsoo Park and Leher Pathak and Carolina Paz and Ludovic Peran and Dmitry Pimenov and Michelle Pokrass and Elizabeth Proehl and Huida Qiu and Gaby Raila and Filippo Raso and Hongyu Ren and Kimmy Richardson and David Robinson and Bob Rotsted and Hadi Salman and Suvansh Sanjeev and Max Schwarzer and D. Sculley and Harshit Sikchi and Kendal Simon and Karan Singhal and Yang Song and Dane Stuckey and Zhiqing Sun and Philippe Tillet and Sam Toizer and Foivos Tsimpourlas and Nikhil Vyas and Eric Wallace and Xin Wang and Miles Wang and Olivia Watkins and Kevin Weil and Amy Wendling and Kevin Whinnery and Cedric Whitney and Hannah Wong and Lin Yang and Yu Yang and Michihiro Yasunaga and Kristen Ying and Wojciech Zaremba and Wenting Zhan and Cyril Zhang and Brian Zhang and Eddie Zhang and Shengjia Zhao},
      year={2025},
      eprint={2508.10925},
      archivePrefix={arXiv},
      primaryClass={cs.CL},
      url={https://arxiv.org/abs/2508.10925}, 
}

@misc{kimiteam2025kimik2openagentic,
      title={Kimi K2{:} Open Agentic Intelligence}, 
      author={Kimi Team and Yifan Bai and Yiping Bao and Guanduo Chen and Jiahao Chen and Ningxin Chen and Ruijue Chen and Yanru Chen and Yuankun Chen and Yutian Chen and Zhuofu Chen and Jialei Cui and Hao Ding and Mengnan Dong and Angang Du and Chenzhuang Du and Dikang Du and Yulun Du and Yu Fan and Yichen Feng and Kelin Fu and Bofei Gao and Hongcheng Gao and Peizhong Gao and Tong Gao and Xinran Gu and Longyu Guan and Haiqing Guo and Jianhang Guo and Hao Hu and Xiaoru Hao and Tianhong He and Weiran He and Wenyang He and Chao Hong and Yangyang Hu and Zhenxing Hu and Weixiao Huang and Zhiqi Huang and Zihao Huang and Tao Jiang and Zhejun Jiang and Xinyi Jin and Yongsheng Kang and Guokun Lai and Cheng Li and Fang Li and Haoyang Li and Ming Li and Wentao Li and Yanhao Li and Yiwei Li and Zhaowei Li and Zheming Li and Hongzhan Lin and Xiaohan Lin and Zongyu Lin and Chengyin Liu and Chenyu Liu and Hongzhang Liu and Jingyuan Liu and Junqi Liu and Liang Liu and Shaowei Liu and T. Y. Liu and Tianwei Liu and Weizhou Liu and Yangyang Liu and Yibo Liu and Yiping Liu and Yue Liu and Zhengying Liu and Enzhe Lu and Lijun Lu and Shengling Ma and Xinyu Ma and Yingwei Ma and Shaoguang Mao and Jie Mei and Xin Men and Yibo Miao and Siyuan Pan and Yebo Peng and Ruoyu Qin and Bowen Qu and Zeyu Shang and Lidong Shi and Shengyuan Shi and Feifan Song and Jianlin Su and Zhengyuan Su and Xinjie Sun and Flood Sung and Heyi Tang and Jiawen Tao and Qifeng Teng and Chensi Wang and Dinglu Wang and Feng Wang and Haiming Wang and Jianzhou Wang and Jiaxing Wang and Jinhong Wang and Shengjie Wang and Shuyi Wang and Yao Wang and Yejie Wang and Yiqin Wang and Yuxin Wang and Yuzhi Wang and Zhaoji Wang and Zhengtao Wang and Zhexu Wang and Chu Wei and Qianqian Wei and Wenhao Wu and Xingzhe Wu and Yuxin Wu and Chenjun Xiao and Xiaotong Xie and Weimin Xiong and Boyu Xu and Jing Xu and Jinjing Xu and L. H. Xu and Lin Xu and Suting Xu and Weixin Xu and Xinran Xu and Yangchuan Xu and Ziyao Xu and Junjie Yan and Yuzi Yan and Xiaofei Yang and Ying Yang and Zhen Yang and Zhilin Yang and Zonghan Yang and Haotian Yao and Xingcheng Yao and Wenjie Ye and Zhuorui Ye and Bohong Yin and Longhui Yu and Enming Yuan and Hongbang Yuan and Mengjie Yuan and Haobing Zhan and Dehao Zhang and Hao Zhang and Wanlu Zhang and Xiaobin Zhang and Yangkun Zhang and Yizhi Zhang and Yongting Zhang and Yu Zhang and Yutao Zhang and Yutong Zhang and Zheng Zhang and Haotian Zhao and Yikai Zhao and Huabin Zheng and Shaojie Zheng and Jianren Zhou and Xinyu Zhou and Zaida Zhou and Zhen Zhu and Weiyu Zhuang and Xinxing Zu},
      year={2025},
      eprint={2507.20534},
      archivePrefix={arXiv},
      primaryClass={cs.LG},
      url={https://arxiv.org/abs/2507.20534}, 
}

@inproceedings{hershcovich-etal-2022-challenges,
    title = "Challenges and Strategies in Cross-Cultural {NLP}",
    author = "Hershcovich, Daniel  and
      Frank, Stella  and
      Lent, Heather  and
      de Lhoneux, Miryam  and
      Abdou, Mostafa  and
      Brandl, Stephanie  and
      Bugliarello, Emanuele  and
      Cabello Piqueras, Laura  and
      Chalkidis, Ilias  and
      Cui, Ruixiang  and
      Fierro, Constanza  and
      Margatina, Katerina  and
      Rust, Phillip  and
      S{\o}gaard, Anders",
    editor = "Muresan, Smaranda  and
      Nakov, Preslav  and
      Villavicencio, Aline",
    booktitle = "Proceedings of the 60th Annual Meeting of the Association for Computational Linguistics (Volume 1: Long Papers)",
    month = may,
    year = "2022",
    address = "Dublin, Ireland",
    publisher = "Association for Computational Linguistics",
    url = "https://aclanthology.org/2022.acl-long.482/",
    doi = "10.18653/v1/2022.acl-long.482",
    pages = "6997--7013"
}

@article{myung2024blend,
  title={Blend: A benchmark for llms on everyday knowledge in diverse cultures and languages},
  author={Myung, Junho and Lee, Nayeon and Zhou, Yi and Jin, Jiho and Putri, Rifki and Antypas, Dimosthenis and Borkakoty, Hsuvas and Kim, Eunsu and Perez-Almendros, Carla and Ayele, Abinew Ali and et al.},
  journal={Advances in Neural Information Processing Systems},
  volume={37},
  pages={78104--78146},
  year={2024}
}

@article{liu-etal-2025-culturally,
    title = "Culturally Aware and Adapted {NLP}: A Taxonomy and a Survey of the State of the Art",
    author = "Liu, Chen Cecilia  and
      Gurevych, Iryna  and
      Korhonen, Anna",
    journal = "Transactions of the Association for Computational Linguistics",
    volume = "13",
    year = "2025",
    address = "Cambridge, MA",
    publisher = "MIT Press",
    url = "https://aclanthology.org/2025.tacl-1.31/",
    doi = "10.1162/tacl_a_00760",
    pages = "652--689"
}

@misc{salemi2023lamp,
      title={La{MP}: When Large Language Models Meet Personalization}, 
      author={Alireza Salemi and Sheshera Mysore and Michael Bendersky and Hamed Zamani},
      year={2023},
      eprint={2304.11406},
      archivePrefix={arXiv},
      primaryClass={cs.CL}
}

@inproceedings{zhao-etal-2024-tracing,
    title = "Tracing the Roots of Facts in Multilingual Language Models: Independent, Shared, and Transferred Knowledge",
    author = "Zhao, Xin  and
      Yoshinaga, Naoki  and
      Oba, Daisuke",
    editor = "Graham, Yvette  and
      Purver, Matthew",
    booktitle = "Proceedings of the 18th Conference of the European Chapter of the Association for Computational Linguistics (Volume 1: Long Papers)",
    month = mar,
    year = "2024",
    address = "St. Julian{'}s, Malta",
    publisher = "Association for Computational Linguistics",
    url = "https://aclanthology.org/2024.eacl-long.127/",
    doi = "10.18653/v1/2024.eacl-long.127",
    pages = "2088--2102"
}

@misc{goldman2025eclekticnovelchallengeset,
      title={ECLeKTic: a Novel Challenge Set for Evaluation of Cross-Lingual Knowledge Transfer}, 
      author={Omer Goldman and Uri Shaham and Dan Malkin and Sivan Eiger and Avinatan Hassidim and Yossi Matias and Joshua Maynez and Adi Mayrav Gilady and Jason Riesa and Shruti Rijhwani and Laura Rimell and Idan Szpektor and Reut Tsarfaty and Matan Eyal},
      year={2025},
      eprint={2502.21228},
      archivePrefix={arXiv},
      primaryClass={cs.CL},
      url={https://arxiv.org/abs/2502.21228}, 
}

@inproceedings{zhang-etal-2025-cross,
    title = "Cross-Lingual Transfer of Cultural Knowledge: An Asymmetric Phenomenon",
    author = "Zhang, Chen  and
      Liao, Zhiyuan  and
      Feng, Yansong",
    editor = "Che, Wanxiang  and
      Nabende, Joyce  and
      Shutova, Ekaterina  and
      Pilehvar, Mohammad Taher",
    booktitle = "Proceedings of the 63rd Annual Meeting of the Association for Computational Linguistics (Volume 2: Short Papers)",
    month = jul,
    year = "2025",
    address = "Vienna, Austria",
    publisher = "Association for Computational Linguistics",
    url = "https://aclanthology.org/2025.acl-short.13/",
    doi = "10.18653/v1/2025.acl-short.13",
    pages = "147--157",
    ISBN = "979-8-89176-252-7"
}

@inproceedings{
li2024culturegen,
title={{CULTURE}-{GEN}: Revealing Global Cultural Perception in Language Models through Natural Language Prompting},
author={Huihan Li and Liwei Jiang and Nouha Dziri and Xiang Ren and Yejin Choi},
booktitle={First Conference on Language Modeling},
year={2024},
url={https://openreview.net/forum?id=DbsLm2KAqP}
}

@inproceedings{singh-etal-2025-global,
    title = "Global {MMLU}: Understanding and Addressing Cultural and Linguistic Biases in Multilingual Evaluation",
    author = "Singh, Shivalika  and
      Romanou, Angelika  and
      Fourrier, Cl{\'e}mentine  and
      Adelani, David Ifeoluwa  and
      Ngui, Jian Gang  and
      Vila-Suero, Daniel  and
      Limkonchotiwat, Peerat  and
      Marchisio, Kelly  and
      Leong, Wei Qi  and
      Susanto, Yosephine  and
      Ng, Raymond  and
      Longpre, Shayne  and
      Ruder, Sebastian  and
      Ko, Wei-Yin  and
      Bosselut, Antoine  and
      Oh, Alice  and
      Martins, Andre  and
      Choshen, Leshem  and
      Ippolito, Daphne  and
      Ferrante, Enzo  and
      Fadaee, Marzieh  and
      Ermis, Beyza  and
      Hooker, Sara",
    editor = "Che, Wanxiang  and
      Nabende, Joyce  and
      Shutova, Ekaterina  and
      Pilehvar, Mohammad Taher",
    booktitle = "Proceedings of the 63rd Annual Meeting of the Association for Computational Linguistics (Volume 1: Long Papers)",
    month = jul,
    year = "2025",
    address = "Vienna, Austria",
    publisher = "Association for Computational Linguistics",
    url = "https://aclanthology.org/2025.acl-long.919/",
    doi = "10.18653/v1/2025.acl-long.919",
    pages = "18761--18799",
    ISBN = "979-8-89176-251-0"
}

@article{lindsey2025biology,
  author={Lindsey, Jack and Gurnee, Wes and Ameisen, Emmanuel and Chen, Brian and Pearce, Adam and Turner, Nicholas L. and Citro, Craig and Abrahams, David and Carter, Shan and Hosmer, Basil and Marcus, Jonathan and Sklar, Michael and Templeton, Adly and Bricken, Trenton and McDougall, Callum and Cunningham, Hoagy and Henighan, Thomas and Jermyn, Adam and Jones, Andy and Persic, Andrew and Qi, Zhenyi and Thompson, T. Ben and Zimmerman, Sam and Rivoire, Kelley and Conerly, Thomas and Olah, Chris and Batson, Joshua},
  title={On the Biology of a Large Language Model},
  journal={Transformer Circuits Thread},
  year={2025},
  url={https://transformer-circuits.pub/2025/attribution-graphs/biology.html}
}

@inproceedings{
zhang2025the,
title={The Same but Different: Structural Similarities and Differences in Multilingual Language Modeling},
author={Ruochen Zhang and Qinan Yu and Matianyu Zang and Carsten Eickhoff and Ellie Pavlick},
booktitle={The Thirteenth International Conference on Learning Representations},
year={2025},
url={https://openreview.net/forum?id=NCrFA7dq8T}
}

@inproceedings{resck_explainability_2025,
  address = {Suzhou, China},
  title = {Explainability and {Interpretability} of {Multilingual} {Large} {Language} {Models}: {A} {Survey}},
  shorttitle = {Explainability and {Interpretability} of {Multilingual} {Large} {Language} {Models}},
  url = {https://openreview.net/forum?id=KQjVhM2YhN},
  language = {en},
  booktitle = {Proceedings of the 2025 {Conference} on {Empirical} {Methods} in {Natural} {Language} {Processing}},
  publisher = {Association for Computational Linguistics},
  author = {Resck, Lucas and Augenstein, Isabelle and Korhonen, Anna},
  month = nov,
  year = {2025},
  bibtex_show = true,
  google_scholar_id = {Zph67rFs4hoC},
  note = {Accepted for publication}
}

@article{ameisen2025circuit,
  author={Ameisen, Emmanuel and Lindsey, Jack and Pearce, Adam and Gurnee, Wes and Turner, Nicholas L. and Chen, Brian and Citro, Craig and Abrahams, David and Carter, Shan and Hosmer, Basil and Marcus, Jonathan and Sklar, Michael and Templeton, Adly and Bricken, Trenton and McDougall, Callum and Cunningham, Hoagy and Henighan, Thomas and Jermyn, Adam and Jones, Andy and Persic, Andrew and Qi, Zhenyi and Ben Thompson, T. and Zimmerman, Sam and Rivoire, Kelley and Conerly, Thomas and Olah, Chris and Batson, Joshua},
  title={Circuit Tracing: Revealing Computational Graphs in Language Models},
  journal={Transformer Circuits Thread},
  year={2025},
  url={https://transformer-circuits.pub/2025/attribution-graphs/methods.html}
}

@inproceedings{
miller2024transformer,
title={Transformer Circuit Evaluation Metrics Are Not Robust},
author={Joseph Miller and Bilal Chughtai and William Saunders},
booktitle={First Conference on Language Modeling},
year={2024},
url={https://openreview.net/forum?id=zSf8PJyQb2}
}

@inproceedings{
zhao2024how,
title={How do Large Language Models Handle Multilingualism?},
author={Yiran Zhao and Wenxuan Zhang and Guizhen Chen and Kenji Kawaguchi and Lidong Bing},
booktitle={The Thirty-eighth Annual Conference on Neural Information Processing Systems},
year={2024},
url={https://openreview.net/forum?id=ctXYOoAgRy}
}

@inproceedings{tang-etal-2024-language,
    title = "Language-Specific Neurons: The Key to Multilingual Capabilities in Large Language Models",
    author = "Tang, Tianyi  and
      Luo, Wenyang  and
      Huang, Haoyang  and
      Zhang, Dongdong  and
      Wang, Xiaolei  and
      Zhao, Xin  and
      Wei, Furu  and
      Wen, Ji-Rong",
    editor = "Ku, Lun-Wei  and
      Martins, Andre  and
      Srikumar, Vivek",
    booktitle = "Proceedings of the 62nd Annual Meeting of the Association for Computational Linguistics (Volume 1: Long Papers)",
    month = aug,
    year = "2024",
    address = "Bangkok, Thailand",
    publisher = "Association for Computational Linguistics",
    url = "https://aclanthology.org/2024.acl-long.309/",
    doi = "10.18653/v1/2024.acl-long.309",
    pages = "5701--5715"
}

@inproceedings{kojima-etal-2024-multilingual,
    title = "On the Multilingual Ability of Decoder-based Pre-trained Language Models: Finding and Controlling Language-Specific Neurons",
    author = "Kojima, Takeshi  and
      Okimura, Itsuki  and
      Iwasawa, Yusuke  and
      Yanaka, Hitomi  and
      Matsuo, Yutaka",
    editor = "Duh, Kevin  and
      Gomez, Helena  and
      Bethard, Steven",
    booktitle = "Proceedings of the 2024 Conference of the North American Chapter of the Association for Computational Linguistics: Human Language Technologies (Volume 1: Long Papers)",
    month = jun,
    year = "2024",
    address = "Mexico City, Mexico",
    publisher = "Association for Computational Linguistics",
    url = "https://aclanthology.org/2024.naacl-long.384/",
    doi = "10.18653/v1/2024.naacl-long.384",
    pages = "6919--6971"
}

@inproceedings{ying-etal-2025-disentangling,
    title = "Disentangling Language and Culture for Evaluating Multilingual Large Language Models",
    author = "Ying, Jiahao  and
      Tang, Wei  and
      Zhao, Yiran  and
      Cao, Yixin  and
      Rong, Yu  and
      Zhang, Wenxuan",
    editor = "Che, Wanxiang  and
      Nabende, Joyce  and
      Shutova, Ekaterina  and
      Pilehvar, Mohammad Taher",
    booktitle = "Proceedings of the 63rd Annual Meeting of the Association for Computational Linguistics (Volume 1: Long Papers)",
    month = jul,
    year = "2025",
    address = "Vienna, Austria",
    publisher = "Association for Computational Linguistics",
    url = "https://aclanthology.org/2025.acl-long.1082/",
    doi = "10.18653/v1/2025.acl-long.1082",
    pages = "22230--22251",
    ISBN = "979-8-89176-251-0"
}

@article{elhage2022superposition,
   title={Toy Models of Superposition},
   author={Elhage, Nelson and Hume, Tristan and Olsson, Catherine and Schiefer, Nicholas and Henighan, Tom and Kravec, Shauna and Hatfield-Dodds, Zac and Lasenby, Robert and Drain, Dawn and Chen, Carol and Grosse, Roger and McCandlish, Sam and Kaplan, Jared and Amodei, Dario and Wattenberg, Martin and Olah, Christopher},
   year={2022},
   journal={Transformer Circuits Thread},
   note={https://transformer-circuits.pub/2022/toy\_model/index.html}
}

@article{bricken2023monosemanticity,
       title={Towards Monosemanticity: Decomposing Language Models With Dictionary Learning},
       author={Bricken, Trenton and Templeton, Adly and Batson, Joshua and Chen, Brian and Jermyn, Adam and Conerly, Tom and Turner, Nick and Anil, Cem and Denison, Carson and Askell, Amanda and Lasenby, Robert and Wu, Yifan and Kravec, Shauna and Schiefer, Nicholas and Maxwell, Tim and Joseph, Nicholas and Hatfield-Dodds, Zac and Tamkin, Alex and Nguyen, Karina and McLean, Brayden and Burke, Josiah E and Hume, Tristan and Carter, Shan and Henighan, Tom and Olah, Christopher},
       year={2023},
       journal={Transformer Circuits Thread},
       note={https://transformer-circuits.pub/2023/monosemantic-features/index.html}
    }

@article{DBLP:journals/corr/abs-2503-05613,
  publtype={informal},
  author={Dong Shu and Xuansheng Wu and Haiyan Zhao and Daking Rai and Ziyu Yao and Ninghao Liu and Mengnan Du},
  title={A Survey on Sparse Autoencoders: Interpreting the Internal Mechanisms of Large Language Models},
  year={2025},
  month={March},
  cdate={1740787200000},
  journal={CoRR},
  volume={abs/2503.05613},
  url={https://doi.org/10.48550/arXiv.2503.05613}
}

@inproceedings{
dunefsky2024transcoders,
title={Transcoders find interpretable {LLM} feature circuits},
author={Jacob Dunefsky and Philippe Chlenski and Neel Nanda},
booktitle={The Thirty-eighth Annual Conference on Neural Information Processing Systems},
year={2024},
url={https://openreview.net/forum?id=J6zHcScAo0}
}

@misc{wiki:Jaccard_index,
   author = "Wikipedia",
   title = "{Jaccard index} --- {W}ikipedia{,} The Free Encyclopedia",
   year = "2025",
   howpublished = {\url{http://en.wikipedia.org/w/index.php?title=Jaccard\%20index&oldid=1292934854}},
   note = "[Online; accessed 14-September-2025]"
 }

@article{DBLP:journals/corr/abs-2408-00118,
  publtype={informal},
  author={Morgane Rivière and Shreya Pathak and Pier Giuseppe Sessa and Cassidy Hardin and Surya Bhupatiraju and Léonard Hussenot and Thomas Mesnard and Bobak Shahriari and Alexandre Ramé and Johan Ferret and Peter Liu and Pouya Tafti and Abe Friesen and Michelle Casbon and Sabela Ramos and Ravin Kumar and Charline Le Lan and Sammy Jerome and Anton Tsitsulin and Nino Vieillard and Piotr Stanczyk and Sertan Girgin and Nikola Momchev and Matt Hoffman and Shantanu Thakoor and Jean-Bastien Grill and Behnam Neyshabur and Olivier Bachem and Alanna Walton and Aliaksei Severyn and Alicia Parrish and Aliya Ahmad and Allen Hutchison and Alvin Abdagic and Amanda Carl and Amy Shen and Andy Brock and Andy Coenen and Anthony Laforge and Antonia Paterson and Ben Bastian and Bilal Piot and Bo Wu and Brandon Royal and Charlie Chen and Chintu Kumar and Chris Perry and Chris Welty and Christopher A. Choquette-Choo and Danila Sinopalnikov and David Weinberger and Dimple Vijaykumar and Dominika Rogozinska and Dustin Herbison and Elisa Bandy and Emma Wang and Eric Noland and Erica Moreira and Evan Senter and Evgenii Eltyshev and Francesco Visin and Gabriel Rasskin and Gary Wei and Glenn Cameron and Gus Martins and Hadi Hashemi and Hanna Klimczak-Plucinska and Harleen Batra and Harsh Dhand and Ivan Nardini and Jacinda Mein and Jack Zhou and James Svensson and Jeff Stanway and Jetha Chan and Jin Peng Zhou and Joana Carrasqueira and Joana Iljazi and Jocelyn Becker and Joe Fernandez and Joost van Amersfoort and Josh Gordon and Josh Lipschultz and Josh Newlan and Ju-yeong Ji and Kareem Mohamed and Kartikeya Badola and Kat Black and Katie Millican and Keelin McDonell and Kelvin Nguyen and Kiranbir Sodhia and Kish Greene and Lars Lowe Sjösund and Lauren Usui and Laurent Sifre and Lena Heuermann and Leticia Lago and Lilly McNealus},
  title={Gemma 2: Improving Open Language Models at a Practical Size},
  year={2024},
  cdate={1704067200000},
  journal={CoRR},
  volume={abs/2408.00118},
  url={https://doi.org/10.48550/arXiv.2408.00118}
}

@inproceedings{lieberum-etal-2024-gemma,
    title = "Gemma Scope: Open Sparse Autoencoders Everywhere All At Once on Gemma 2",
    author = "Lieberum, Tom  and
      Rajamanoharan, Senthooran  and
      Conmy, Arthur  and
      Smith, Lewis  and
      Sonnerat, Nicolas  and
      Varma, Vikrant  and
      Kramar, Janos  and
      Dragan, Anca  and
      Shah, Rohin  and
      Nanda, Neel",
    editor = "Belinkov, Yonatan  and
      Kim, Najoung  and
      Jumelet, Jaap  and
      Mohebbi, Hosein  and
      Mueller, Aaron  and
      Chen, Hanjie",
    booktitle = "Proceedings of the 7th BlackboxNLP Workshop: Analyzing and Interpreting Neural Networks for NLP",
    month = nov,
    year = "2024",
    address = "Miami, Florida, US",
    publisher = "Association for Computational Linguistics",
    url = "https://aclanthology.org/2024.blackboxnlp-1.19/",
    doi = "10.18653/v1/2024.blackboxnlp-1.19",
    pages = "278--300"
}

@inproceedings{mitchell-etal-2022-enhancing,
    title = "Enhancing Self-Consistency and Performance of Pre-Trained Language Models through Natural Language Inference",
    author = "Mitchell, Eric  and
      Noh, Joseph  and
      Li, Siyan  and
      Armstrong, Will  and
      Agarwal, Ananth  and
      Liu, Patrick  and
      Finn, Chelsea  and
      Manning, Christopher",
    editor = "Goldberg, Yoav  and
      Kozareva, Zornitsa  and
      Zhang, Yue",
    booktitle = "Proceedings of the 2022 Conference on Empirical Methods in Natural Language Processing",
    month = dec,
    year = "2022",
    address = "Abu Dhabi, United Arab Emirates",
    publisher = "Association for Computational Linguistics",
    url = "https://aclanthology.org/2022.emnlp-main.115/",
    doi = "10.18653/v1/2022.emnlp-main.115",
    pages = "1754--1768"
}

\end{document}